\documentclass[runningheads]{llncs}

\usepackage{eccv}

\usepackage{eccvabbrv}
\usepackage{graphicx}
\usepackage{booktabs}
\usepackage{amsmath,amssymb}
\usepackage{xcolor}
\usepackage{siunitx}
\usepackage{hyperref}
\definecolor{linknavy}{RGB}{20,60,140}
\hypersetup{colorlinks=true, urlcolor=linknavy, linkcolor=linknavy,
            citecolor=linknavy, breaklinks=true}
\usepackage{orcidlink}

\definecolor{ourgreen}{RGB}{46,158,91}
\definecolor{ourred}{RGB}{200,45,45}
\newcommand{\good}[1]{\textcolor{ourgreen}{#1}}
\newcommand{\bad}[1]{\textcolor{ourred}{#1}}
\newcommand{\apd}{\mathrm{APD}}
\newcommand{\best}[1]{\textbf{#1}}
\usepackage{colortbl}
\definecolor{winrow}{RGB}{237,247,240}
\definecolor{barred}{RGB}{223,138,138}
\definecolor{bargreen}{RGB}{113,181,124}
\newcommand{\dbar}[2]{%
  \makebox[27pt][r]{\scriptsize\textcolor{barred}{$#2$}}\,%
  \textcolor{barred}{\rule[-0.4pt]{#1pt}{4.6pt}}\makebox[0pt][l]{}}
\newcommand{\gbar}[2]{%
  \makebox[27pt][r]{\scriptsize\textcolor{bargreen}{$#2$}}\,%
  \textcolor{bargreen}{\rule[-0.4pt]{#1pt}{4.6pt}}\makebox[0pt][l]{}}
\newcommand{\second}[1]{\underline{#1}}

\definecolor{ruleboxbg}{RGB}{246,248,250}
\definecolor{ruleboxln}{RGB}{94,143,196}
\makeatletter
\newsavebox{\g@rulebox}
\newenvironment{rulebox}{%
  \par\addvspace{1.4ex}%
  \begin{lrbox}{\g@rulebox}%
  \begin{minipage}{\dimexpr\linewidth-1.75em\relax}%
  \small\itshape\setlength{\parskip}{0pt}%
}{%
  \end{minipage}\end{lrbox}%
  \noindent\begingroup\setlength{\fboxsep}{0.6em}%
  \colorbox{ruleboxbg}{\textcolor{ruleboxln}{%
    \rule[-\dp\g@rulebox]{2.2pt}{\dimexpr\ht\g@rulebox+\dp\g@rulebox\relax}}%
    \hspace{0.5em}\usebox{\g@rulebox}}%
  \endgroup\par\addvspace{1.4ex}%
}
\makeatother

\begin{document}

\title{Gallileo-4D: Frozen Backbone Ensemble for\\Dynamic 4D Reconstruction}
\titlerunning{Gallileo-4D: Frozen Backbone Ensemble}

\author{Nicol\`o Savioli}
\authorrunning{N.~Savioli}
\institute{OdaxAI Research\\[2pt]
\href{https://odaxai.com}{odaxai.com}\\[1pt]
\email{nicolo.savioli@odaxai.com}}

\maketitle

\begin{center}\vspace{-2.2ex}
{\small
\href{https://github.com/odaxai/Gallileo-4D}{%
  \raisebox{-0.28\height}{\includegraphics[height=1.05em]{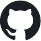}}\;Code}%
\hspace{2.2em}%
\href{https://huggingface.co/OdaxAI/gallileo-4d-weights}{%
  \raisebox{-0.28\height}{\includegraphics[height=1.15em]{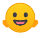}}\;Weights}%
}
\end{center}
\vspace{0.4ex}

\begin{abstract}
We describe our entry to the PhysAI Dynamic 4D Reconstruction Challenge, which
placed third of $27$ teams at $0.58356$ APD on the final leaderboard, without a
single gradient update. This was not the plan: of thirteen fine-tuning
configurations of a pre-trained 4D backbone, twelve degraded the challenge
score, and eleven of those twelve improved local validation at the same time.
We trace this inversion to the structure of the benchmark: only $25\%$ of the
evaluation set belongs to the data variant released for training, so updates
that fit the available data damage the pre-trained features the remaining
$75\%$ relies on. Our system therefore freezes the backbone and spends its
budget at inference time, fusing three decoding configurations---temporal
stride-3, horizontal-flip test-time augmentation, and dense stride-1---under a
convex weighting. The ensemble recovers $+0.041$ APD over the frozen baseline,
more than any training run achieved, at zero training cost.

\keywords{4D reconstruction \and Point tracking \and Distribution shift
\and Test-time ensembling \and Frozen features}
\end{abstract}

\begin{figure}[t]
  \centering
  \includegraphics[width=0.67\linewidth]{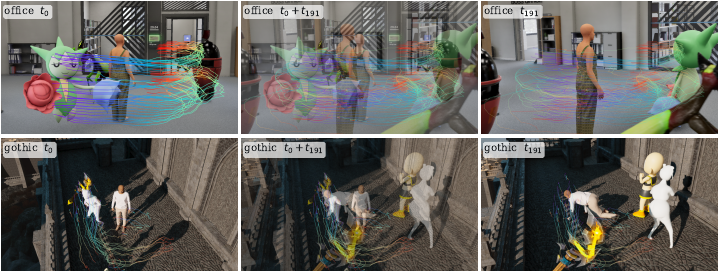}
  \caption{\textbf{Gallileo-4D results.} Dynamic 4D reconstruction of two
  challenge test sequences, \emph{office} (top) and \emph{gothic} (bottom).
  Each row shows the reconstruction at the first queried timestamp $t_0$, an
  $\alpha$-blend of the first and last, and the last timestamp $t_{191}$, with
  the recovered 3D trajectory of every tracked point drawn over the full
  sequence and coloured by time. The system placed third of $27$ teams on the final
  leaderboard \emph{without a single gradient update}. A uniform tone curve is
  applied to the low-light gothic panels for print legibility; no geometry or
  trajectory is retouched.}
  \label{fig:teaser}
\end{figure}
\section{Introduction}
\label{sec:intro}

Reconstructing the 3D geometry of a scene together with its evolution over
time from a single monocular video---4D reconstruction---has moved rapidly
from per-scene optimisation towards feed-forward
inference~\cite{feng2025st4rtrack,karhade2025any4d,sucar2026vdpm,luo2026arc}.
The PhysAI Dynamic 4D Reconstruction Challenge evaluates this capability on
$128$ held-out synthetic sequences of $192$ frames each, scoring the predicted
world-space position of $512$ query points at $32$ timestamps per sequence.

We entered the challenge with a conventional plan: take the strongest
available feed-forward backbone, fine-tune it on the released training split
with a differentiable surrogate of the evaluation metric, and submit. The plan failed in an unusually clean way. Across thirteen training
configurations---spanning learning rate, epoch count, loss formulation, and
which part of the network receives gradients---exactly one improved the
challenge score, by $+0.028$: less than what we later obtained by averaging
two decoding passes. The other twelve regressed, several catastrophically.
Worse, the regressions were invisible locally: our held-out validation score
kept improving while the challenge score fell (Fig.~\ref{fig:failures}a).

Rather than treat this as a tuning problem, we treated it as evidence.
Section~\ref{sec:analysis} shows that the two facts have a common cause. The
only training data released to us covers one of the four data variants present
in the test set; scene identities are disjoint; clips are $50$ frames rather
than $192$ and square rather than widescreen. A local validation set drawn
from that training split therefore measures performance on a distribution that
accounts for a quarter of the score, and fine-tuning on it is a textbook
setting for feature distortion under distribution
shift~\cite{kumar2022finetuning}.

The resulting system is deliberately unambitious in its learning and
aggressive in its inference. We keep the backbone frozen and combine three decoding strategies under a
convex weighting selected by a two-parameter grid search. We name it Gallileo-4D, after a countryman who is remembered for
writing down what the instrument showed rather than what the theory required.

\smallskip
\noindent\textbf{Contributions.}
\begin{itemize}
  \setlength{\itemsep}{5pt}
  \setlength{\parskip}{0pt}
  \setlength{\parsep}{0pt}
  \setlength{\topsep}{6pt}
  \setlength{\leftmargin}{1em}
  \item A documented negative result: thirteen fine-tuning configurations of a
  state-of-the-art 4D backbone, twelve of which regress on the challenge
  metric, reported with per-run learning rates, losses and scores
  (Sec.~\ref{sec:failures}, Tab.~\ref{tab:failed}).
  \item An analysis of \emph{why} local validation inverts in this setting,
  including an axis that gained $+0.0129$ locally and lost $-0.3443$ on the
  leaderboard, and the operating rule we derived from it
  (Sec.~\ref{sec:analysis}).
  \item Gallileo-4D, a frozen-backbone inference-time ensemble that reaches
  third place with no weight updates, together with its full ablation
  (Sec.~\ref{sec:method},~\ref{sec:experiments}).
  \item Characterisation of a scale-collapse failure mode in which raising the
  inference resolution from $512$ to $1008$\,px drives the predicted scene
  scale $30\times$ below its expected value (Sec.~\ref{sec:scale}).
\end{itemize}
\smallskip

\section{Related Work}
\label{sec:related}

\smallskip
\noindent\textbf{Feed-forward 3D and 4D reconstruction.}
DUSt3R~\cite{wang2024dust3r} and MASt3R~\cite{leroy2024mast3r} established that
dense stereo geometry can be regressed in a single forward pass;
VGGT~\cite{wang2025vggt}, Pi3~\cite{wang2025pi3} and Depth Anything
3~\cite{lin2025da3} scale this to many views. Dynamic extensions differ mainly
in how motion is represented, from per-frame geometry on dynamic
input~\cite{zhang2025monst3r} and paired point maps~\cite{feng2025st4rtrack} to
B\'ezier trajectory fields~\cite{liu2025trace}, first-frame-relative
motion~\cite{karhade2025any4d} and dynamic point maps~\cite{sucar2026vdpm};
per-scene optimisation~\cite{mildenhall2020nerf,kerbl20233dgs,wang2024shapeofmotion}
is accurate but far too slow for a $128$-sequence benchmark. Our backbone is
4RC~\cite{luo2026arc}, used strictly as a fixed function. The challenge metric
is closest to 3D point-tracking benchmarks such as
TAPVid-3D~\cite{koppula2024tapvid3d} and inherits its median-scale alignment;
tracking-specific systems~\cite{karaev2023cotracker,xiao2024spatialtracker} are
strong on sparse queries but do not produce the dense geometry the challenge
also scores.

\smallskip
\noindent\textbf{Fine-tuning under shift, ensembles, adaptive evaluation.}
That fine-tuning can underperform the frozen features it starts from is
documented outside reconstruction: it distorts pre-trained features and degrades
out-of-distribution accuracy relative to linear probing~\cite{kumar2022finetuning},
which WiSE-FT~\cite{wortsman2022wiseft} repairs by interpolating back towards
the zero-shot weights. The classical framing is catastrophic
interference~\cite{mccloskey1989catastrophic}, mitigated by
EWC~\cite{kirkpatrick2017ewc} or PackNet~\cite{mallya2018packnet}; we find a
low-rank restriction of the update~\cite{hu2022lora} does not prevent the effect
here (Sec.~\ref{sec:failures}). On the aggregation side, averaging independent
predictions reduces variance~\cite{lakshminarayanan2017ensembles} and test-time
augmentation applies the idea to input transformations~\cite{shanmugam2021tta};
weight averaging~\cite{wortsman2022soups} needs several trained checkpoints,
which our findings make undesirable, so we ensemble in output space over
decoding configurations of a \emph{single} frozen network. Any selection on a
public split is adaptive data
analysis~\cite{dwork2015holdout,recht2019imagenet}; we keep ours
two-dimensional and quantify the exposure in Sec.~\ref{sec:discussion}.

\section{Analysis of the Challenge}
\label{sec:analysis}

\subsection{Setup and metric}
\label{sec:setup}

The challenge is built on the Syn4D benchmark~\cite{syn4d2026}. The evaluation
set contains $128$ sequences of $192$ frames at $1280\times720$, drawn from
four scene families (antiquity, dream, gothic, office) and four rendering
variants (\texttt{og}, \texttt{sim}, \texttt{mixed},
\texttt{mixed\_no\_bedlam}), $32$ sequences per variant. For each sequence,
$512$ query points defined on a source frame must be localised in world
coordinates at $32$ target timestamps, giving $2{,}097{,}152$ predicted 3D
positions per submission.

Scoring uses the Average Percentage of points within Distance. Predictions are
first aligned to the ground truth by a per-sequence median scale factor
\begin{equation}
  s \;=\; \operatorname{median}_{q,t}
  \frac{\lVert P^{*}_{q,t}\rVert_2}{\lVert \hat{P}_{q,t}\rVert_2},
  \label{eq:scale}
\end{equation}
after which the fraction of points falling inside a threshold ball is averaged
over four thresholds $\delta \in \{0.1, 0.3, 0.5, 1.0\}$\,m:
\begin{equation}
  \apd_{\mathcal{S}} =
  \frac{1}{4}\sum_{\delta}
  \frac{1}{|\mathcal{S}|}\sum_{(q,t)\in\mathcal{S}}
  \mathbf{1}\!\left[\, \lVert s\,\hat{P}_{q,t} - P^{*}_{q,t}\rVert_2 < \delta \,\right].
  \label{eq:apd}
\end{equation}
The reported score balances all queries against the moving subset,
$\apd = \tfrac{1}{2}(\apd_{\text{all}} + \apd_{\text{dyn}})$, making it far
more sensitive to the small fraction of genuinely dynamic points than a uniform
average would be. We reimplemented Eqs.~\eqref{eq:scale}--\eqref{eq:apd} and
verified them against the organisers' reference implementation with zero
deviation. Because Eq.~\eqref{eq:scale} removes any global scale error, the
metric is invariant to the overall size of the reconstruction but extremely
sensitive to its \emph{internal} consistency---a property that matters in
Sec.~\ref{sec:scale}.

\subsection{A training split covering a quarter of the test set}
\label{sec:domaingap}

\begin{figure}[t]
  \centering
  \includegraphics[width=0.76\linewidth]{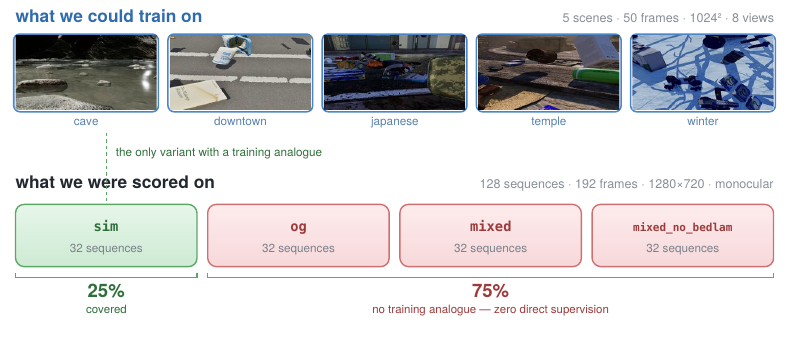}
  \caption{\textbf{What we could train on versus what we were scored on.}
  Top: the five scenes of the released \texttt{syn4d\_sim} split, the only
  training data available to us. Middle: the evaluation set is divided evenly
  across four rendering variants, and only \texttt{sim} has a training
  analogue, leaving three quarters of the distribution without direct
  supervision. Bottom: the three calibration points we obtained between the
  local proxy and the challenge score, two of which invert in sign. The
  SlotTABA bar is truncated at the plot edge; its true value is $-0.3443$.}
  \label{fig:domaingap}
\end{figure}

\begin{table}[t]
\centering
\footnotesize
\setlength{\tabcolsep}{3.2pt}
\begin{tabular}{@{}lcc@{}}
\toprule
& \multicolumn{2}{c}{\textbf{Split}} \\
\cmidrule(l){2-3}
\textbf{Attribute} & \texttt{syn4d\_sim} (train) & challenge (test) \\
\midrule
Frames / clip      & 50                       & 192 \\
Resolution         & $1024{\times}1024$       & $1280{\times}720$ \\
Aspect ratio       & $1{:}1$                  & $16{:}9$ \\
Cameras / clip     & 8 (multi-view)           & 1 (monocular) \\
Horiz.\ FOV        & $35$--$84^\circ$          & $39.8$--$89.7^\circ$ \\
Scene families     & 5                        & 4 \\
Scene overlap      & \multicolumn{2}{c}{$0\%$, disjoint names} \\
Rendering variants & \bad{$1$ (\texttt{sim})} & $4$, uniform \\
\midrule
Test coverage      & \multicolumn{2}{c}{\bad{$25\%$ of test has a train analogue}} \\
\bottomrule
\end{tabular}
\caption{Training split versus evaluation set. The variant row determines
everything else in this paper.}
\label{tab:domaingap}
\end{table}

Table~\ref{tab:domaingap} compares the released training split with the
evaluation set. Several axes differ, but only one is decisive. The training
archive contains a single rendering variant, \texttt{syn4d\_sim}; the test set
contains four in equal proportion. We verified this: the released tree exposes only
\texttt{syn4d\_sim/} and \texttt{sim/}, and no additional splits exist on the
competition data tab, the participant kit, or the dataset host.

Concretely, $32$ of $128$ evaluation sequences ($25\%$) are in-distribution
with respect to content, and $96$ ($75\%$) require cross-variant
generalisation for which zero direct supervision exists. Any parameter update
that improves the \texttt{sim} quarter at the expense of general geometric
priors is therefore a net loss in expectation, and it is a loss that a
\texttt{sim}-only validation set cannot see.

Two secondary axes constrain what our proxy can test. Training clips are $50$
frames; test sequences are $192$. Any inference-time question about temporal
windowing---window length, stride, overlap---is thus unanswerable locally,
since a $96$-frame window does not fit in a $50$-frame clip. Field of view, by
contrast, turned out \emph{not} to be a gap. We list it because it was our
leading hypothesis before measurement, and it was wrong.

\subsection{Local validation inverts}
\label{sec:proxy}

\begin{table}[t]
\centering
\footnotesize
\setlength{\tabcolsep}{4.5pt}
\begin{tabular}{@{}lccl@{}}
\toprule
& \multicolumn{2}{c}{\textbf{Score change}} & \\
\cmidrule(lr){2-3}
\textbf{Intervention} & local & leaderb. & \textbf{Verdict} \\
\midrule
base $\rightarrow$ geo ckpt      & $+0.0078$ & \good{$+0.0325$} & agrees, $4.2\times$ \\
geo $\rightarrow$ SlotTABA       & $+0.0129$ & \bad{$-0.3443$}  & false positive \\
geo $\rightarrow$ fine-tune      & $+0.035$  & \bad{$-0.030$}   & false positive \\
\bottomrule
\end{tabular}
\caption{Every calibration point obtained between the local proxy and the
challenge leaderboard. Two of three are sign-inconsistent; the fine-tuning row
is the mean over the twelve regressing runs of Tab.~\ref{tab:failed}.}
\label{tab:proxy}
\end{table}

We built a local proxy from $31$ held-out \texttt{syn4d\_sim} sequences with
reconstructed ground truth, and measured its resolution before trusting it.
Recovering a query's 3D position two ways---direct depth unprojection versus
face-identifier centroid---gives a median disagreement of $4.3$\,mm and a
$95$th percentile of $1.6$\,cm, comfortably inside the finest $0.1$\,m
threshold. However, face-identifier persistence across frames is weak: only
$16$--$57\%$ of (query, frame) cells match successfully after the first frame,
with a mean of ${\sim}25\%$. The effective sample size for the dynamic half of
the metric is therefore about a quarter of nominal, and we treat any local
delta below $0.005$ $\apd$ as noise.

Table~\ref{tab:proxy} lists every calibration point we obtained. The single
agreeing point, a checkpoint swap, showed a leaderboard gain $4.2\times$
larger than the local gain---already enough to make the proxy useless as a
magnitude estimator. The two disagreeing points are worse. A slot-based
temporal aggregation module raised the local score by $+0.0129$ and dropped
the leaderboard score from $0.512$ to $0.177$, a loss of $-0.3443$; it had
been accepted locally because the acceptance test scored bearing consistency,
which the module optimised, while silently destroying the depth structure the
metric depends on.

From three points we adopted the only rule they support, and followed it for
the rest of the competition:

\begin{rulebox}
\textbf{Operating rule.} Local validation is a \emph{regression veto}, never a
promotion gate. A configuration that scores worse locally is blocked. A
configuration that scores equal or better becomes \emph{eligible} for
evaluation on the public split; it is never assumed to be an improvement.
\end{rulebox}

This rule is cheap to state and expensive to follow: it converts the local
proxy from a decision procedure into a filter, and defers final judgement to
the public split on the held-out score. We
believe it is nonetheless the correct rule whenever the validation split
covers a minority of the evaluation distribution.

\section{What Did Not Work}
\label{sec:failures}

\begin{figure}[t]
  \centering
  \includegraphics[width=0.81\linewidth]{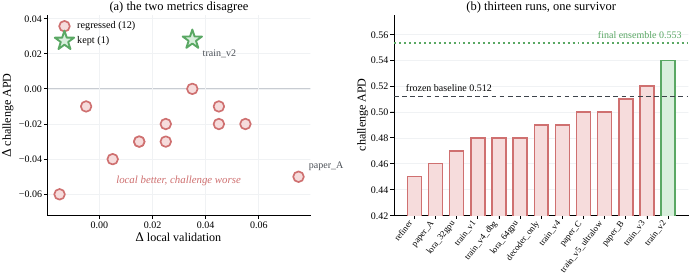}
  \caption{\textbf{Thirteen training runs, one survivor.} (a) Change in local
  validation against change in challenge $\apd$, both relative to the frozen
  baseline. Eleven runs land above the horizontal axis on the local proxy and
  below it on the challenge metric. Only \texttt{train\_v2} reaches the
  positive quadrant. (b) Absolute challenge score per run, annotated with
  learning rate; the dashed line is the frozen baseline the runs were supposed
  to beat, and the dotted line is the final ensemble, which uses no training
  at all.}
  \label{fig:failures}
\end{figure}

This section is the part of our submission we consider most useful to other
participants, because it is the part that consumed most of the compute.
Table~\ref{tab:failed} lists all thirteen runs. Training used the challenge
metric's differentiable surrogate,
\begin{equation}
  \mathcal{L}_{\text{soft}} = 1 - \frac{1}{|\mathcal{T}|}
  \sum_{\delta}\sum_{i}
  \sigma\!\left(\frac{\delta - \lVert \hat{P}^{\text{aln}}_{i} - P^{*}_{i}\rVert_2}{\tau}\right),
  \label{eq:softapd}
\end{equation}
with $\tau = 0.05$ and $\hat{P}^{\text{aln}}$ the scale-aligned prediction.
Because Eq.~\eqref{eq:scale} makes the scale itself a function of the
prediction, backpropagating through it is unstable; we therefore detach it,
\begin{equation}
  s = \frac{\langle \mathrm{sg}(\hat{P}), P^{*}\rangle}{\lVert \mathrm{sg}(\hat{P})\rVert_2^{2}},
  \qquad
  \hat{P}^{\text{aln}} = \hat{P}\cdot \mathrm{sg}(s),
  \label{eq:scaledec}
\end{equation}
with $\mathrm{sg}(\cdot)$ the stop-gradient. This scale-decoupled formulation
did stabilise optimisation. It did not improve the leaderboard.

\begin{table}[t]
\centering
\footnotesize
\setlength{\tabcolsep}{4.2pt}
\renewcommand{\arraystretch}{1.12}
\begin{tabular}{@{}llcc cc c@{}}
\toprule
\textbf{Run} & \textbf{Adapted part} & \textbf{LR} & \textbf{ep.} &
\textbf{local} & \textbf{chall.} & \textbf{$\Delta$ vs.\ frozen} \\
\midrule
\addlinespace[1pt]
\multicolumn{7}{@{}l}{\itshape Na\"ive fine-tuning}\\[1pt]
\texttt{train\_v1}  & full network     & \num{5e-5} & 1 & 0.42 & 0.48 & \dbar{13.4}{-0.032} \\
\rowcolor{winrow}
\texttt{train\_v2}  & full network     & \num{5e-6} & 1 & 0.44 & \best{0.54} & \gbar{11.7}{+0.028} \\
\texttt{train\_v3}  & full network     & \num{5e-6} & 3 & 0.46 & \second{0.52} & \gbar{3.4}{+0.008} \\
\texttt{train\_v4}  & full, scale-dec. & \num{1e-5} & 1 & 0.45 & 0.49 & \dbar{9.2}{-0.022} \\
\texttt{train\_v4d} & full, scale-dec. & \num{1e-5} & 1 & 0.43 & 0.48 & \dbar{13.4}{-0.032} \\
\texttt{train\_v5}  & full network     & \num{1e-7} & 1 & 0.40 & 0.50 & \dbar{5.0}{-0.012} \\
\addlinespace[2pt]\midrule\addlinespace[1pt]
\multicolumn{7}{@{}l}{\itshape Aggressive schedules}\\[1pt]
\texttt{paper\_A}   & full network     & \num{1e-4} & 5 & \underline{0.48} & 0.46 & \dbar{21.8}{-0.052} \\
\texttt{paper\_B}   & full network     & \num{1e-6} & 2 & 0.44 & \second{0.51} & \dbar{0.8}{-0.002} \\
\texttt{paper\_C}   & full, cosine     & \num{5e-6} & 2 & 0.45 & 0.50 & \dbar{5.0}{-0.012} \\
\addlinespace[2pt]\midrule\addlinespace[1pt]
\multicolumn{7}{@{}l}{\itshape Parameter-efficient adaptation}\\[1pt]
\texttt{lora\_32}   & LoRA, rank $16$  & \num{1e-4} & 1 & 0.41 & 0.47 & \dbar{17.6}{-0.042} \\
\texttt{lora\_64}   & LoRA, rank $32$  & \num{5e-5} & 1 & 0.42 & 0.48 & \dbar{13.4}{-0.032} \\
\addlinespace[2pt]\midrule\addlinespace[1pt]
\multicolumn{7}{@{}l}{\itshape Lightweight refinement}\\[1pt]
\texttt{refiner}    & added MLP head   & \num{1e-3} & 1 & 0.39 & 0.45 & \dbar{26.0}{-0.062} \\
\texttt{decoder}    & decoder only     & \num{5e-5} & 1 & 0.43 & 0.49 & \dbar{9.2}{-0.022} \\
\addlinespace[2pt]\midrule\addlinespace[1pt]
\rowcolor{winrow}
\textbf{none} & \textbf{frozen $+$ ensemble} & --- & --- & --- & \best{0.553} & \gbar{17.2}{+0.041} \\
\bottomrule
\end{tabular}
\caption{All thirteen training runs, with the frozen inference-time ensemble in
the last row for reference. Bars are drawn to scale; green is a gain over the
frozen \texttt{geofinetune} baseline at $0.512$ APD, red a loss. Only
\texttt{train\_v2} and \texttt{train\_v3} gained, and only \texttt{train\_v2}
by more than noise --- while \texttt{paper\_A} took the best \underline{local}
score of the study. Local scores are soft-APD on the $31$-sequence
\texttt{sim} proxy and are not comparable in absolute value to the challenge
column.}
\label{tab:failed}
\end{table}

\smallskip
\noindent\textbf{Na\"ive fine-tuning.} Learning rate behaves monotonically and
in the wrong direction: \num{5e-5} costs $-0.032$, \num{5e-6} gains $+0.028$,
and \num{1e-7} costs $-0.012$ by learning nothing while still perturbing batch
statistics. Extending \texttt{train\_v2} from one epoch to three
(\texttt{train\_v3}) raises the local score from $0.44$ to $0.46$ and lowers
the challenge score from $0.54$ to $0.52$. This is the cleanest single
instance of the inversion: the additional epochs measurably fit the
\texttt{sim} variant and measurably damage the rest.

\smallskip
\noindent\textbf{Aggressive schedules.} \texttt{paper\_A}, at \num{1e-4} for
five epochs, achieves the best local score of any run ($0.48$) and the
second-worst challenge score ($0.46$). If one had been selecting on local
validation, this is the checkpoint one would have shipped.

\smallskip
\noindent\textbf{Parameter-efficient adaptation.} We expected LoRA to help,
since restricting the update to a low-rank subspace is a standard defence
against forgetting. It did not: rank $16$ at \num{1e-4} costs $-0.042$ and
rank $32$ at \num{5e-5} costs $-0.032$, both worse than full fine-tuning at a
comparable learning rate. We read this as evidence that the problem is not the
\emph{capacity} of the update but its \emph{objective}---a low-rank direction
that reduces loss on \texttt{sim} is still a direction away from the features
the other three variants need.

\smallskip
\noindent\textbf{Lightweight refinement.} Freezing the backbone and training
only an added MLP refinement head produced the single worst result of the
study ($-0.062$). Training only the decoder with a frozen encoder cost
$-0.022$. Both confirm that the damage does not require touching the encoder.

\subsection{Scale collapse}
\label{sec:scale}

\begin{figure}[tb]
  \centering
  \includegraphics[width=0.65\linewidth]{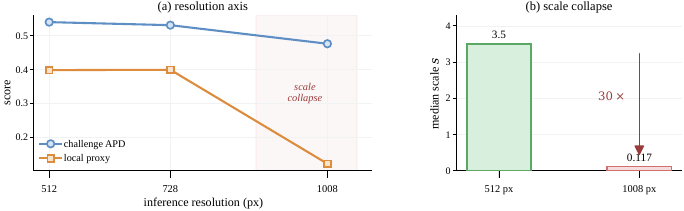}
  \caption{\textbf{The resolution axis and its failure mode.} (a) Local and
  challenge scores against inference resolution. (b) At $1008$\,px the
  geometry head emits predictions whose median scale is $0.117$ against an
  expected $3.5$, a collapse that median alignment cannot repair because it is
  not globally consistent.}
  \label{fig:scale}
\end{figure}

One axis failed for a mechanically different and instructive reason. Raising
the inference resolution is normally a safe way to gain accuracy. Here it
destroys the prediction: at $1008$\,px the local score falls from $0.398$ to
$0.121$ and the challenge score from $0.540$ to $0.476$. The cause is visible
in the raw outputs. The median per-sequence scale factor of
Eq.~\eqref{eq:scale} moves from $3.5$ at $512$\,px to $0.117$ at $1008$\,px
(Fig.~\ref{fig:scale}b), \ie{} the geometry head emits depths roughly
$30\times$ smaller than expected. The checkpoint was never trained at that
resolution---the backbone's own protocol samples long edges up to
$504$\,px---and the metric's global median alignment cannot rescue it, because
the collapse is not a single global factor but varies across the scene. Adding
$1008$\,px predictions to our ensemble lowered the final score from $0.55345$
to $0.54923$, so we excluded the axis entirely.

\section{Method: Gallileo-4D}
\label{sec:method}

\begin{figure}[t]
  \centering
  \includegraphics[width=0.81\linewidth]{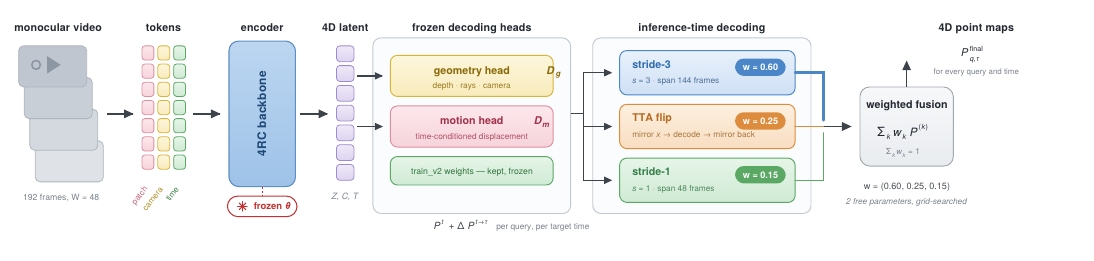}
  \caption{\textbf{Model architecture of Gallileo-4D.} The video is
  patchified into per-frame patch, camera and time tokens and encoded once by
  the frozen 4RC backbone into the 4D latent $\mathcal{F}$. Frozen geometry
  and motion heads decode a base point map plus a displacement for any (query,
  target time) pair; the same frozen network is evaluated under three
  inference-time decoding configurations, and their point maps are averaged
  with fixed weights. No parameter is updated at any stage: the only fitted
  quantities in the entire system are the three fusion weights $w$.}
  \label{fig:arch}
\end{figure}

\subsection{Base model}

We use 4RC~\cite{luo2026arc} with the public \texttt{geofinetune} checkpoint
as $f_\theta$, keeping $\theta$ fixed. Its encoder is a DINOv2-initialised
ViT-G~\cite{oquab2023dinov2} pre-trained on a mixture of dynamic and static
corpora including PointOdyssey~\cite{zheng2023pointodyssey} and
Kubric~\cite{greff2022kubric}, none of which overlaps the challenge scenes.
Given a window of frames $\mathcal{V} = \{I_i\}$, the backbone encodes once
and decodes the position of query $q$ at target time $\tau$ as a base geometry
plus a displacement,
\begin{equation}
  \hat{P}^{t_q\rightarrow\tau}_{q} =
  \hat{P}^{t_q}_{q} + \Delta\hat{P}^{t_q\rightarrow\tau}_{q}.
\end{equation}
The $192$-frame sequences are processed with a sliding window of $W=48$ frames
at $512$\,px, the resolution the checkpoint was trained for. Prediction heads
from \texttt{train\_v2}, the one surviving run of Sec.~\ref{sec:failures}, are
used in place of the released heads; these are also frozen.
Figure~\ref{fig:arch} shows the resulting system.

\subsection{Why freeze}

Three properties of this challenge make a frozen backbone the right default
rather than a fallback. \emph{(i)} The training distribution covers $25\%$ of
the evaluation distribution (Sec.~\ref{sec:domaingap}), so the expected value
of a gradient step is negative once the damage to the uncovered $75\%$ is
priced in. \emph{(ii)} Local validation cannot detect that damage
(Sec.~\ref{sec:proxy}), so there is no mechanism to stop early.
\emph{(iii)} Empirically, twelve of thirteen attempts regressed
(Sec.~\ref{sec:failures}). Freezing converts an unbounded and unobservable
risk into a fixed, known baseline of $0.512$, and lets the remaining budget be
spent on interventions whose effect is directly measurable on the evaluation of record.

\subsection{Decoding configurations}

\begin{figure}[t]
  \centering
  \includegraphics[width=0.70\linewidth]{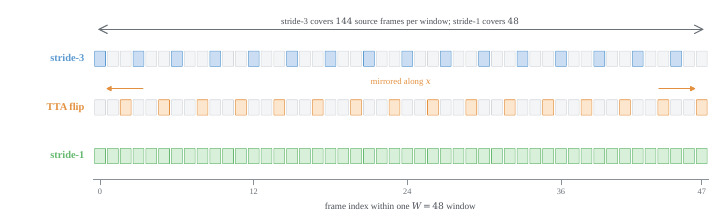}\\[3pt]
  \includegraphics[width=0.76\linewidth]{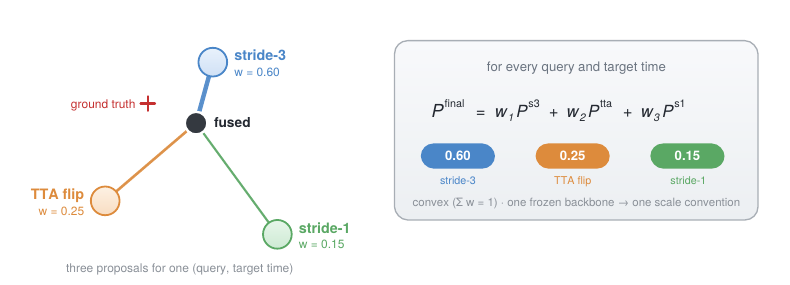}
  \caption{\textbf{The three decoding configurations, and how they combine.}
  Top: which frames each configuration admits inside one $W{=}48$ window.
  Stride-3 admits every third frame and so reaches $144$ source frames of
  temporal context; stride-1 admits all $48$ and maximises temporal resolution
  instead; the TTA branch runs the stride-3 pattern on a horizontally mirrored
  window and mirrors the predicted point maps back. The three therefore make
  errors that are only weakly correlated, which is what the fusion exploits.
  Bottom: for each query $q$ and target time $\tau$ the three configurations
  propose three world-space positions, and the submitted prediction is their
  fixed convex combination. Averaging in world coordinates is meaningful only
  because all three share the frozen backbone, and therefore one scale
  convention.}
  \label{fig:sampling}
\end{figure}

We evaluate the same frozen network under three decoding configurations chosen
to make different errors rather than better ones
(Fig.~\ref{fig:sampling}).

\smallskip
\noindent\textbf{Stride-3} ($w_1 = 0.60$). Every third frame enters the
window, so a single window spans $144$ source frames. Long temporal context is
the regime in which the backbone's global attention is most useful, and this
configuration is the strongest individually ($0.543$).

\smallskip
\noindent\textbf{Horizontal-flip TTA} ($w_2 = 0.25$). The window is mirrored,
decoded, and the resulting point maps are mirrored back,
$\hat{P}^{(2)} = \mathcal{F}^{-1}(f_\theta(\mathcal{F}(\mathcal{V})))$, with
$\mathcal{F}$ the horizontal flip acting on the $x$ axis of the world frame.
This axis was \emph{blocked} by our local proxy, which showed a $-0.0021$
regression with $18$ of $31$ sequences worse; under the veto rule of
Sec.~\ref{sec:proxy} a regression blocks promotion, and we submitted it only
after the frozen-backbone reading of Sec.~\ref{sec:analysis} made us
re-examine axes the proxy had rejected. It gains $+0.003$.

\smallskip
\noindent\textbf{Stride-1} ($w_3 = 0.15$). Every frame enters the window,
maximising temporal resolution at the cost of context length. Alone it is
weaker than stride-3, but its errors on fast motion are largely uncorrelated,
which is what the ensemble exploits: its marginal contribution ($+0.009$) is
three times that of TTA despite the smaller weight.

\subsection{Fusion and weight optimisation}

\begin{figure}[t]
  \centering
  \includegraphics[width=0.81\linewidth]{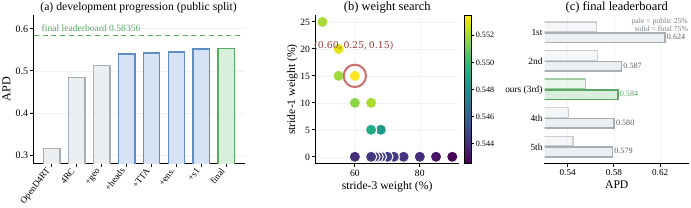}
  \caption{\textbf{Results.} (a) Public-split score after each accepted change during
  development, from the organisers' OpenD4RT baseline to our submitted system;
  grey bars are public baselines, blue are inference-time changes, green is the
  submitted entry, and the dashed line is the score that entry finally
  received on the private split. (b) The ensemble weight search of Eq.~\eqref{eq:fusion}; the optimum is
  circled and the surface around it is flat. (c) Final leaderboard: pale bars
  the public $25\%$, solid the final $75\%$; other entries unnamed.}
  \label{fig:results}
\end{figure}

The three predictions are combined per query and timestamp
(Fig.~\ref{fig:sampling}),
\begin{equation}
  \hat{P}^{\text{final}}_{q,\tau} =
  w_1 \hat{P}^{\text{s3}}_{q,\tau} +
  w_2 \hat{P}^{\text{tta}}_{q,\tau} +
  w_3 \hat{P}^{\text{s1}}_{q,\tau},
  \quad \textstyle\sum_k w_k = 1 .
  \label{eq:fusion}
\end{equation}
The simplex has two free parameters. Because the local proxy cannot rank
configurations (Sec.~\ref{sec:proxy}), we selected them with a small
coarse-to-fine grid evaluated on the public split, first along the $w_3 = 0$
edge, then over the interior once stride-1 predictions were available. The
optimum sits at $(0.60, 0.25, 0.15)$ and, importantly, the surface is flat
around it: the eight best configurations lie within $0.0015$ $\apd$ of each
other (Fig.~\ref{fig:results}b). The result is therefore insensitive to the
exact triple, and Sec.~\ref{sec:discussion} bounds the residual selection
effect.

\section{Experiments}
\label{sec:experiments}

\smallskip
\noindent\textbf{Setup.} Training consumed roughly $1{,}000$ GPU-hours and
produced one usable artefact; inference consumed $168$ GPU-hours and produced
the submission.

\subsection{Main results}

\begin{table}[t]
\centering
\footnotesize
\setlength{\tabcolsep}{7pt}
\begin{tabular}{@{}rlccc@{}}
\toprule
& & \multicolumn{2}{c}{\textbf{APD} $\uparrow$} & \\
\cmidrule(lr){3-4}
\textbf{Rank} & \textbf{Entry} & public & \textbf{final} & \textbf{$\Delta$} \\
\midrule
1 & anonymised            & 0.56496 & \best{0.62426} & $+0.0593$ \\
2 & anonymised            & 0.56557 & 0.58667 & $+0.0211$ \\
\rowcolor{winrow}
3 & \textbf{OdaxAI (ours)} & 0.55513 & \second{0.58356} & $+0.0284$ \\
4 & anonymised            & 0.54082 & 0.58024 & $+0.0394$ \\
5 & anonymised            & 0.54455 & 0.57876 & $+0.0342$ \\
\midrule
--- & 4RC public submission         & 0.48419 & 0.51142 & $+0.0272$ \\
--- & organisers' OpenD4RT baseline & 0.31564 & 0.32217 & $+0.0065$ \\
\bottomrule
\end{tabular}
\caption{Final leaderboard, top five. The public column is scored on
approximately $25\%$ of the test data and the final column on the remaining
$75\%$. Identities are withheld; only our own entry is named. Our submitted
system moved up one place between the two, and the configuration analysed in
Sec.~\ref{sec:method} and ablated in Tab.~\ref{tab:ablation} is the one
scored here.}
\label{tab:leaderboard}
\end{table}

Table~\ref{tab:leaderboard} gives the final standings alongside the public
ones, and the comparison is the most informative result in this paper. Every
entry scored higher on the private $75\%$ than on the public $25\%$, so the
two splits are not interchangeable in absolute terms. What matters is the
\emph{relative} movement: our entry gained $+0.0284$ and rose one place, while
the entries immediately below us on the public split gained $+0.0394$ and
$+0.0342$ and stayed below. The system did not degrade when the evaluation set
grew fourfold.

This is a direct test of the prediction we made in Sec.~\ref{sec:discussion}
before the private split was released: that the ensemble gain would survive
because the weight surface is flat and the selection was two-dimensional. It
did. We note the converse honestly --- the first-place entry gained $+0.0593$,
twice our movement, which is the signature of a method that was \emph{under}
selected on the public split rather than over-selected on it. Whatever they
did, they left more on the table publicly than we did, and it was not
leaderboard fitting.

\subsection{Ablation}

\begin{table}[t]
\centering
\footnotesize
\setlength{\tabcolsep}{5pt}
\begin{tabular}{@{}lccccl@{}}
\toprule
& \multicolumn{3}{c}{\textbf{Fusion weights}} & & \\
\cmidrule(lr){2-4}
\textbf{Configuration} & $w_1$ & $w_2$ & $w_3$ & \textbf{$\apd$} $\uparrow$ & \\
\midrule
OpenD4RT baseline             & ---  & ---  & ---  & 0.31564 & \\
4RC, released ckpt            & ---  & ---  & ---  & 0.48419 & \\
$+$ \texttt{geofinetune}      & ---  & ---  & ---  & 0.51234 & \\
$+$ \texttt{train\_v2} heads  & ---  & ---  & ---  & 0.54001 & \\
\midrule
stride-3 only                 & 1.00 & ---  & ---  & 0.54256 & \\
$+$ TTA                       & 0.70 & 0.30 & ---  & 0.54452 & \\
$+$ TTA                       & 0.68 & 0.32 & ---  & 0.54455 & \\
$+$ TTA $+$ stride-1          & 0.68 & 0.27 & 0.05 & 0.54846 & \\
$+$ TTA $+$ stride-1          & 0.65 & 0.25 & 0.10 & 0.55206 & \\
\textbf{Gallileo-4D}          & 0.60 & 0.25 & 0.15 & \best{0.55345} & \\
$+$ TTA $+$ stride-1          & 0.55 & 0.25 & 0.20 & \second{0.55298} & \\
$+$ TTA $+$ stride-1          & 0.50 & 0.25 & 0.25 & 0.55201 & \\
\midrule
$+$ res-$1008$ branch         & \multicolumn{3}{c}{4-way} & \bad{0.54923} & \\
\bottomrule
\end{tabular}
\caption{Ablation of the ensemble. All rows use the same frozen backbone; only
the decoding configurations and their weights change. \best{Best} and
\second{second best}. The four-way row adds the collapsed $1008$\,px branch of
Sec.~\ref{sec:scale} and was rejected.}
\label{tab:ablation}
\end{table}

Table~\ref{tab:ablation} decomposes the final score. The frozen pipeline
reaches $0.54001$ before any ensembling. Fusion then adds $+0.01344$, which is
larger than the contribution of every training run in Tab.~\ref{tab:failed}
combined, and was obtained for the cost of two extra forward passes per
sequence.

The internal structure of that gain is worth noting. Along the two-component
edge the score saturates quickly---$w_1$ from $0.90$ to $0.68$ buys only
$+0.002$, and the optimum at $0.68/0.32$ is barely distinguishable from
$0.70/0.30$. Introducing the third component breaks the saturation: $5\%$ of
stride-1 weight adds $+0.004$, $10\%$ adds $+0.0072$, and $15\%$ adds
$+0.0089$, after which the curve turns over. Diversity of temporal sampling,
not the quality of any individual pass, is what the ensemble is buying.

\subsection{Qualitative results}

\begin{figure}[tb]
  \centering
  \includegraphics[width=0.70\linewidth]{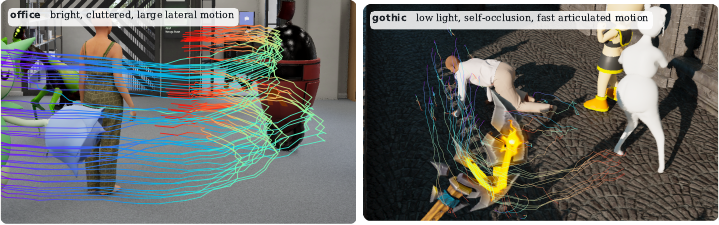}
  \caption{\textbf{Reconstruction detail on two test sequences.} Left,
  \emph{office}: a bright cluttered interior with large lateral motion, where
  trajectories stay smooth and separated even as actors cross. Right,
  \emph{gothic}: low light, self-occlusion and fast articulated motion;
  bundles stay coherent through the occlusion, but tracks on the fastest limb
  fan out, which is where the residual error concentrates.}
  \label{fig:qual}
\end{figure}

Figure~\ref{fig:teaser} shows two challenge test sequences at the first and
last queried timestamps, and Fig.~\ref{fig:qual} shows the same
reconstructions at higher magnification. The frozen backbone recovers coherent static geometry
and temporally smooth trajectories for the dynamic content in both the bright
and the low-light regime. Residual artefacts are of two kinds. Trajectories
belonging to the fastest-moving limbs fan out near their endpoints, which is
consistent with the limitations reported for the backbone on chaotic motion.
And thin planar slivers appear in the reconstructed background of the office
scene, the signature of the base geometry losing depth resolution at distance
on low-texture surfaces; we leave both visible rather than crop them away.

\begin{figure}[htb]
  \centering
  \includegraphics[width=0.63\linewidth]{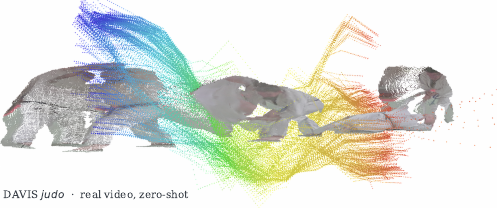}
  \caption{\textbf{Zero-shot transfer to real video.} The frozen pipeline
  applied unchanged to the \emph{judo} sequence of
  DAVIS~\cite{perazzi2016davis}: reconstructed geometry with the dense 3D
  motion field of the throw, coloured by time. Neither the decoding heads nor
  the fusion weights were touched, and the system was never trained or tuned
  on real footage. The figure is qualitative only---DAVIS provides no metric
  4D ground truth---but it shows that nothing in the recipe is specific to the
  synthetic benchmark.}
  \label{fig:davis}
\end{figure}

\smallskip
\noindent\textbf{Beyond the benchmark.} The system contains no weights fitted
to this challenge beyond three fusion scalars, so it should transfer to data it
has never seen. Figure~\ref{fig:davis} applies the pipeline unchanged to a
real video from DAVIS~\cite{perazzi2016davis}: the recovered motion field
follows the articulated throw through contact and partial occlusion. We show
one sequence and claim nothing quantitative from it.

\section{Discussion}
\label{sec:discussion}

\smallskip
\noindent\textbf{Why fine-tuning fails here.} The mechanism matches the
account of Kumar~\etal~\cite{kumar2022finetuning}: gradient descent on a
narrow slice of the target distribution moves features away from the
configuration supporting the rest of it, while the loss on that slice falls
throughout. The challenge supplies an unusually stark version---the covered
slice is exactly $25\%$, the uncovered variants are rendered differently
rather than merely sampled differently, and the only validation data available
comes from the covered slice. We would therefore expect the inversion of
Fig.~\ref{fig:failures}a to be reproducible rather than incidental. A
WiSE-FT-style interpolation~\cite{wortsman2022wiseft} towards the released
weights is the obvious untried remedy.

\smallskip
\noindent\textbf{The same critique applies to us.} The public leaderboard on
which we selected $w$ is itself computed on approximately $25\%$ of the test
data. The symmetry with our central finding is exact, and we would rather state
it than let a reader discover it: the argument we make against our local proxy
applies, in weaker form, to the signal we replaced it with.

Two things make the weaker form genuinely weaker. The public split is a
\emph{random} subset of the same distribution, whereas our local proxy was a
different rendering variant---a covariate shift, not a sampling error. And the
selection is two-dimensional on a simplex, far below the dimensionality at
which public-split overfitting is typically
observed~\cite{recht2019imagenet,dwork2015holdout}, with a flat objective: the
eight best configurations span $0.0015$ $\apd$. We therefore expected the ensemble gain of
$+0.041$ to survive on the private split and the third decimal of the specific
triple not to; Sec.~\ref{sec:experiments} reports that it did.

\smallskip
\noindent\textbf{Cost and limitations.} The system runs the backbone three
times per sequence, so a submission costs about $3\times$ a single pass;
acceptable for an offline benchmark, not for deployment, where the honest
comparison is the $0.54001$ single-pass baseline rather than the top of the
leaderboard. Our
conclusions come from one benchmark with one backbone, and the weights are
selected on one public split. The $1{,}000$ GPU-hours were not an exhaustive
search: we did not try weight-space interpolation, variant-conditional
adaptation, or synthesising the missing variants. We report that fine-tuning
failed for us under thirteen configurations, not that it cannot succeed.

\section{Conclusions}
\label{sec:conclusion}

We placed third of $27$ teams on the final leaderboard of the PhysAI Dynamic
4D Reconstruction Challenge with a system containing no weights of our own.
Twelve of thirteen fine-tuning configurations degraded the challenge score
while improving local validation, because the released training split covers a
quarter of the evaluation distribution and the local proxy is drawn from that
same quarter. Freezing the backbone and spending the budget on inference-time
diversity recovered $+0.041$ $\apd$, more than any training run produced. The
lesson is about measurement rather than architecture: a validation signal
covering a minority of the evaluation distribution stops estimating performance
and starts estimating overfitting to that minority, and the two are hard to tell
apart from inside --- a caution that applies to the public leaderboard as much
as to our local proxy. Code and logs: \url{https://github.com/odaxai/Gallileo-4D}.

\smallskip\noindent\textbf{Acknowledgements.} We thank the PhysAI organisers
for the Syn4D benchmark and the 4RC authors for their checkpoints.

\bibliographystyle{splncs04}
\bibliography{refs}

@article{luo2026arc,
  title={4RC: 4D Reconstruction via Conditional Querying Anytime and Anywhere},
  author={Luo, Yihang and Zhou, Shangchen and Lan, Yushi and Pan, Xingang and Loy, Chen Change},
  journal={arXiv preprint arXiv:2602.10094},
  year={2026},
  doi={10.48550/arXiv.2602.10094}
}

@article{sucar2026vdpm,
  title={V-DPM: 4D Video Reconstruction with Dynamic Point Maps},
  author={Sucar, Edgar and Insafutdinov, Eldar and Lai, Zihang and Vedaldi, Andrea},
  journal={arXiv preprint arXiv:2601.09499},
  year={2026},
  doi={10.48550/arXiv.2601.09499}
}

@article{karhade2025any4d,
  title={Any4D: Unified Feed-Forward Metric 4D Reconstruction},
  author={Karhade, Jay and Keetha, Nikhil and Zhang, Yuchen and Gupta, Tanisha and Sharma, Akash and Scherer, Sebastian and Ramanan, Deva},
  journal={arXiv preprint arXiv:2512.10935},
  year={2025},
  doi={10.48550/arXiv.2512.10935}
}

@article{syn4d2026,
  title={Syn4D: A Multiview Synthetic 4D Dataset},
  author={Jiang, Zeren and Lan, Yushi and Luo, Yihang and Deng, Yufan and Lai, Zihang and Sucar, Edgar and Rupprecht, Christian and Laina, Iro and Larlus, Diane and Zheng, Chuanxia and Vedaldi, Andrea},
  journal={arXiv preprint arXiv:2605.05207},
  year={2026},
  doi={10.48550/arXiv.2605.05207}
}

@inproceedings{wang2024dust3r,
  title={{DUSt3R}: Geometric 3D Vision Made Easy},
  author={Wang, Shuzhe and Leroy, Vincent and Cabon, Yohann and Chidlovskii, Boris and Revaud, Jerome},
  booktitle={CVPR},
  year={2024},
  doi={10.48550/arXiv.2312.14132}
}

@inproceedings{leroy2024mast3r,
  title={Grounding Image Matching in 3D with {MASt3R}},
  author={Leroy, Vincent and Cabon, Yohann and Revaud, Jerome},
  booktitle={ECCV},
  year={2024},
  doi={10.1007/978-3-031-73220-1_5}
}

@inproceedings{wang2025vggt,
  title={{VGGT}: Visual Geometry Grounded Transformer},
  author={Wang, Jianyuan and Chen, Minghao and Karaev, Nikita and Vedaldi, Andrea and Rupprecht, Christian and Novotny, David},
  booktitle={CVPR},
  year={2025},
  doi={10.48550/arXiv.2503.11651}
}

@inproceedings{zhang2025monst3r,
  title={{MonST3R}: A Simple Approach for Estimating Geometry in the Presence of Motion},
  author={Zhang, Junyi and Herrmann, Charles and Hur, Junhwa and Jampani, Varun and Darrell, Trevor and Cole, Forrester and Sun, Deqing and Yang, Ming-Hsuan},
  booktitle={ICLR},
  year={2025},
  doi={10.48550/arXiv.2410.03825}
}

@article{wang2025pi3,
  title={Pi3: Permutation-Equivariant Visual Geometry Learning},
  author={Wang, Yifan and Zhou, Jianjun and Zhu, Haoyi and Chang, Wenzheng and Zhou, Yang and Li, Zizun and Chen, Junyi and Pang, Jiangmiao and Shen, Chunhua and He, Tong},
  journal={arXiv preprint arXiv:2507.13347},
  year={2025},
  doi={10.48550/arXiv.2507.13347}
}

@article{lin2025da3,
  title={Depth Anything 3: Recovering the Visual Space from Any Views},
  author={Lin, Haotong and Chen, Sili and Liew, Jun Hao and Chen, Donny Y. and Li, Zhenyu and Shi, Guang and Feng, Jiashi and Kang, Bingyi},
  journal={arXiv preprint arXiv:2511.10647},
  year={2025},
  doi={10.48550/arXiv.2511.10647}
}

@inproceedings{feng2025st4rtrack,
  title={{St4RTrack}: Simultaneous 4D Reconstruction and Tracking in the World},
  author={Feng, Haiwen and Zhang, Junyi and Wang, Qianqian and Ye, Yufei and Yu, Pengcheng and Black, Michael J. and Darrell, Trevor and Kanazawa, Angjoo},
  booktitle={ICCV},
  year={2025},
  doi={10.48550/arXiv.2504.13152}
}

@article{liu2025trace,
  title={Trace Anything: Representing Any Video in 4D via Trajectory Fields},
  author={Liu, Xinhang and Xiao, Yuxi and Chen, Donny Y. and Feng, Jiashi and Tai, Yu-Wing and Tang, Chi-Keung and Kang, Bingyi},
  journal={arXiv preprint},
  year={2025},
  doi={10.48550/arXiv.2510.13802}
}

@inproceedings{xiao2024spatialtracker,
  title={{SpatialTracker}: Tracking Any 2D Pixels in 3D Space},
  author={Xiao, Yuxi and Wang, Qianqian and Zhang, Shangzhan and Xue, Nan and Peng, Sida and Shen, Yujun and Zhou, Xiaowei},
  booktitle={CVPR},
  year={2024},
  doi={10.48550/arXiv.2404.04319}
}

@inproceedings{karaev2023cotracker,
  title={{CoTracker}: It Is Better to Track Together},
  author={Karaev, Nikita and Rocco, Ignacio and Graham, Benjamin and Neverova, Natalia and Vedaldi, Andrea and Rupprecht, Christian},
  booktitle={ECCV},
  year={2024},
  doi={10.48550/arXiv.2307.07635}
}

@inproceedings{mildenhall2020nerf,
  title={{NeRF}: Representing Scenes as Neural Radiance Fields for View Synthesis},
  author={Mildenhall, Ben and Srinivasan, Pratul P. and Tancik, Matthew and Barron, Jonathan T. and Ramamoorthi, Ravi and Ng, Ren},
  booktitle={ECCV},
  year={2020},
  doi={10.1007/978-3-030-58452-8_24}
}

@article{kerbl20233dgs,
  title={3D Gaussian Splatting for Real-Time Radiance Field Rendering},
  author={Kerbl, Bernhard and Kopanas, Georgios and Leimk{\"u}hler, Thomas and Drettakis, George},
  journal={ACM Transactions on Graphics},
  volume={42},
  number={4},
  year={2023},
  doi={10.1145/3592433}
}

@article{wang2024shapeofmotion,
  title={Shape of Motion: 4D Reconstruction from a Single Video},
  author={Wang, Qianqian and Ye, Vickie and Gao, Hang and Austin, Jake and Li, Zhengqi and Kanazawa, Angjoo},
  journal={arXiv preprint arXiv:2407.13764},
  year={2024},
  doi={10.48550/arXiv.2407.13764}
}

@incollection{mccloskey1989catastrophic,
  title={Catastrophic Interference in Connectionist Networks: The Sequential Learning Problem},
  author={McCloskey, Michael and Cohen, Neal J.},
  booktitle={Psychology of Learning and Motivation},
  volume={24},
  pages={109--165},
  year={1989},
  publisher={Academic Press},
  doi={10.1016/S0079-7421(08)60536-8}
}

@article{kirkpatrick2017ewc,
  title={Overcoming Catastrophic Forgetting in Neural Networks},
  author={Kirkpatrick, James and Pascanu, Razvan and Rabinowitz, Neil and Veness, Joel and Desjardins, Guillaume and Rusu, Andrei A. and Milan, Kieran and Quan, John and Ramalho, Tiago and Grabska-Barwinska, Agnieszka and others},
  journal={Proceedings of the National Academy of Sciences},
  volume={114},
  number={13},
  pages={3521--3526},
  year={2017},
  doi={10.1073/pnas.1611835114}
}

@inproceedings{mallya2018packnet,
  title={{PackNet}: Adding Multiple Tasks to a Single Network by Iterative Pruning},
  author={Mallya, Arun and Lazebnik, Svetlana},
  booktitle={CVPR},
  year={2018},
  doi={10.1109/CVPR.2018.00810}
}

@inproceedings{kumar2022finetuning,
  title={Fine-Tuning Can Distort Pretrained Features and Underperform Out-of-Distribution},
  author={Kumar, Ananya and Raghunathan, Aditi and Jones, Robbie and Ma, Tengyu and Liang, Percy},
  booktitle={ICLR},
  year={2022},
  doi={10.48550/arXiv.2202.10054}
}

@inproceedings{wortsman2022wiseft,
  title={Robust Fine-Tuning of Zero-Shot Models},
  author={Wortsman, Mitchell and Ilharco, Gabriel and Kim, Jong Wook and Li, Mike and Kornblith, Simon and Roelofs, Rebecca and Lopes, Raphael Gontijo and Hajishirzi, Hannaneh and Farhadi, Ali and Namkoong, Hongseok and Schmidt, Ludwig},
  booktitle={CVPR},
  year={2022},
  doi={10.48550/arXiv.2109.01903}
}

@inproceedings{wortsman2022soups,
  title={Model Soups: Averaging Weights of Multiple Fine-Tuned Models Improves Accuracy Without Increasing Inference Time},
  author={Wortsman, Mitchell and Ilharco, Gabriel and Gadre, Samir Yitzhak and Roelofs, Rebecca and Gontijo-Lopes, Raphael and Morcos, Ari S. and Namkoong, Hongseok and Farhadi, Ali and Carmon, Yair and Kornblith, Simon and Schmidt, Ludwig},
  booktitle={ICML},
  year={2022},
  doi={10.48550/arXiv.2203.05482}
}

@inproceedings{hu2022lora,
  title={{LoRA}: Low-Rank Adaptation of Large Language Models},
  author={Hu, Edward J. and Shen, Yelong and Wallis, Phillip and Allen-Zhu, Zeyuan and Li, Yuanzhi and Wang, Shean and Wang, Lu and Chen, Weizhu},
  booktitle={ICLR},
  year={2022},
  doi={10.48550/arXiv.2106.09685}
}

@inproceedings{lakshminarayanan2017ensembles,
  title={Simple and Scalable Predictive Uncertainty Estimation Using Deep Ensembles},
  author={Lakshminarayanan, Balaji and Pritzel, Alexander and Blundell, Charles},
  booktitle={NeurIPS},
  year={2017},
  doi={10.48550/arXiv.1612.01474}
}

@inproceedings{shanmugam2021tta,
  title={Better Aggregation in Test-Time Augmentation},
  author={Shanmugam, Divya and Blalock, Davis and Balakrishnan, Guha and Guttag, John},
  booktitle={ICCV},
  year={2021},
  doi={10.1109/ICCV48922.2021.00125}
}

@article{oquab2023dinov2,
  title={{DINOv2}: Learning Robust Visual Features Without Supervision},
  author={Oquab, Maxime and Darcet, Timoth{\'e}e and Moutakanni, Th{\'e}o and Vo, Huy V. and Szafraniec, Marc and Khalidov, Vasil and others},
  journal={arXiv preprint arXiv:2304.07193},
  year={2023},
  doi={10.48550/arXiv.2304.07193}
}

@inproceedings{zheng2023pointodyssey,
  title={{PointOdyssey}: A Large-Scale Synthetic Dataset for Long-Term Point Tracking},
  author={Zheng, Yang and Harley, Adam W. and Shen, Bokui and Wetzstein, Gordon and Guibas, Leonidas J.},
  booktitle={ICCV},
  year={2023},
  doi={10.48550/arXiv.2307.15055}
}

@inproceedings{greff2022kubric,
  title={Kubric: A Scalable Dataset Generator},
  author={Greff, Klaus and Belletti, Francois and Beyer, Lucas and Doersch, Carl and Du, Yilun and others},
  booktitle={CVPR},
  year={2022},
  doi={10.1109/CVPR52688.2022.00373}
}

@inproceedings{koppula2024tapvid3d,
  title={{TAPVid-3D}: A Benchmark for Tracking Any Point in 3D},
  author={Koppula, Skanda and Rocco, Ignacio and Yang, Yi and Heyward, Joe and Carreira, Jo{\~a}o and Zisserman, Andrew and Brostow, Gabriel and Doersch, Carl},
  booktitle={NeurIPS},
  year={2024},
  doi={10.48550/arXiv.2407.05921}
}

@inproceedings{recht2019imagenet,
  title={Do {ImageNet} Classifiers Generalize to {ImageNet}?},
  author={Recht, Benjamin and Roelofs, Rebecca and Schmidt, Ludwig and Shankar, Vaishaal},
  booktitle={ICML},
  year={2019},
  doi={10.48550/arXiv.1902.10811}
}

@article{dwork2015holdout,
  title={The Reusable Holdout: Preserving Validity in Adaptive Data Analysis},
  author={Dwork, Cynthia and Feldman, Vitaly and Hardt, Moritz and Pitassi, Toniann and Reingold, Omer and Roth, Aaron},
  journal={Science},
  volume={349},
  number={6248},
  pages={636--638},
  year={2015},
  doi={10.1126/science.aaa9375}
}

@inproceedings{perazzi2016davis,
  title={A Benchmark Dataset and Evaluation Methodology for Video Object Segmentation},
  author={Perazzi, Federico and Pont-Tuset, Jordi and McWilliams, Brian and Van Gool, Luc and Gross, Markus and Sorkine-Hornung, Alexander},
  booktitle={CVPR},
  year={2016},
  doi={10.1109/CVPR.2016.85}
}

\end{document}